\documentclass[10pt,twocolumn,letterpaper]{article}

\usepackage[pagenumbers]{cvpr}      
\definecolor{cvprblue}{rgb}{0.21,0.49,0.74}
\usepackage[pagebackref,breaklinks,colorlinks,allcolors=cvprblue]{hyperref}
\usepackage{hyperref}
\usepackage{graphicx}
\graphicspath{{./images/}}
\usepackage{multirow}

\usepackage{algorithm}
\usepackage{alphalph}
\usepackage{algorithmic}
\usepackage{bbding}
\usepackage{makecell}
\usepackage[accsupp]{axessibility}

\usepackage{orcidlink}

\usepackage{enumitem}

\usepackage{subcaption}

\usepackage[utf8]{inputenc} 
\usepackage[T1]{fontenc}    
\usepackage{url}            
\usepackage{booktabs}       
\usepackage{amsfonts}       
\usepackage{nicefrac}       
\usepackage{microtype}      
\usepackage{amssymb}
\usepackage{pifont}

\usepackage[table,dvipsnames]{xcolor}
\definecolor{LightCyan}{rgb}{0.91,0.91,0.98}
\definecolor{LightYellow}{rgb}{1.0, 1.0, 0.88}
\definecolor{magicmint}{rgb}{0.67, 0.94, 0.82}
\definecolor{lightmauve}{rgb}{0.86, 0.82, 1.0}
\definecolor{grannysmithapple}{rgb}{0.66, 0.89, 0.63}
\definecolor{isabelline}{rgb}{0.95,0.93,0.91}

\def\confName{CVPR}
\def\confYear{2026}

\title{AgentRVOS for MeViS-Text Track of 5th PVUW Challenge: 3rd Method}
\author{
Deshui Miao$^{1}$ \quad
Chao Yang$^{1}$ \quad
Chao Tian$^{1}$ \\
Guoqing Zhu$^{1}$ \quad
Kai Yang$^{2}$ \quad
Zhifan Mo$^{3}$ \quad
Xin Li$^{1*}$ \\
$^{1}$Pengcheng Laboratory \quad
$^{2}$Wuhan Textile University \\
$^{3}$Yixiang Innovation Technology (Shenzhen) \\
}

\begin{document}
\maketitle
\begin{abstract}
This report describes a Ref-VOS pipeline centered on Sa2VA and organized with explicit agent roles. The key idea is that Sa2VA should provide the first dense semantic hypothesis, while an agent loop decides whether that hypothesis should be accepted, revised, or refined. The pipeline starts with a target-presence judgment stage. If the referred object does not exist in the video, the system directly outputs zero masks. Otherwise, Sa2VA receives the video and referring prompt and produces a coarse mask trajectory over the full video. This trajectory is treated as a semantic prior rather than a final answer. A planner agent decomposes the query, temporal partition agents identify informative blocks, scout agents search for anchor frames, and refinement agents convert reliable Sa2VA masks into boxes and points for SAM3 propagation. A critic scores candidate trajectories, a reflection controller repairs weak hypotheses, and a collaboration controller reconciles multiple agent branches. The result is a Ref-VOS system in which Sa2VA is responsible for dense grounded understanding, while the agent layer handles presence verification, temporal search, confidence-aware revision, and final mask refinement.
\end{abstract}

\section{Introduction}
Referring video object segmentation (Ref-VOS)~\cite{ding2023mevis, li2021referring, vltpami} aims to predict frame-wise pixel masks for the object specified by a natural-language expression~\cite{cheng2022xmem,wu2022language,gong2025reinforcing,videolisa,zheng2024villa}. Although recent foundation models have substantially improved language-conditioned video understanding, solving Ref-VOS in realistic settings remains challenging~\cite{hyperseg,luo2024soc,liu2024universal}. In particular, the task often cannot be reliably completed by a single end-to-end model invocation~\cite{ding2023mevis,tian2025dtos,videolisa}. 
CVPR 2026 5th PVUW challenge has three tracks: Complex VOS on
MOSEv2~\cite{ding2025mosev2}, targeting realistic, cluttered scenes with
small, occluded, reappearing, and camouflaged objects
under adverse conditions; VOS on MOSE~\cite{ding2023mose}, focusing
on challenging, long and diverse videos; and RVOS on
MeViS~\cite{ding2023mevis, MeViSv2}, assessing referring video object segmentation with text or audio.
Even a strong model such as Sa2VA~\cite{yuan2025sa2va} may exhibit two recurring failure modes. First, it can be compelled to return a mask for a target that is in fact absent from the video. Second, when the referred object is small, visually ambiguous, or only partially visible across time, the predicted masks may become coarse, temporally unstable, or poorly localized. These errors are not merely failures of mask prediction; more fundamentally, they reflect unresolved decisions about target existence, temporal validity, anchor-frame reliability, and prediction confidence.

Motivated by this observation, we adopt a design in which Sa2VA~\cite{yuan2025sa2va} serves as the semantic backbone, while the overall Ref-VOS process is governed by an agentic control loop. The role of Sa2VA is to provide an initial language-grounded interpretation of the query over the video and to produce coarse mask hypotheses together with temporal evidence. However, this initial prediction is not treated as universally trustworthy. Instead, the system explicitly reasons about whether the referred target exists, identifies where the initial interpretation is reliable or uncertain, and determines how subsequent refinement should be carried out.

This formulation naturally yields a division of labor between semantic grounding and decision-making~\cite{zhang2024ferret,wang2025time,bai2025univg,yang2023set}. Sa2VA~\cite{yuan2025sa2va}  is responsible for coarse semantic localization and preliminary mask generation, whereas the agent layer performs higher-level control: it interprets the query, selects informative frames, evaluates the credibility of Sa2VA~\cite{yuan2025sa2va} outputs, and decides when stronger geometric refinement is necessary. For the latter, the agent invokes SAM3~\cite{carion2025sam}  as a dedicated mask refinement and propagation module to improve spatial precision and temporal consistency.

\begin{figure*}
    \centering
    \resizebox{\linewidth}{!}{
        \includegraphics{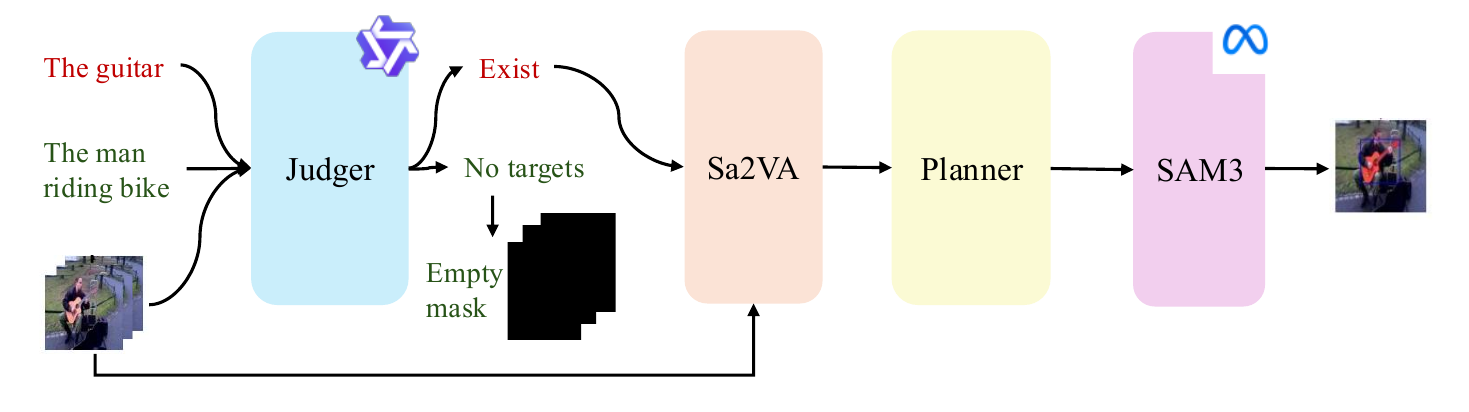}
    }
    \caption{\textbf{Pipeline of our methods.}}
    \label{fig:mevis_audio}
     \vspace{-4mm}
\end{figure*}

\section{Problem Formulation}
Given an input video $V=\{I_t\}_{t=1}^{T}$ and a natural-language referring expression $q$, the goal of referring video object segmentation is to predict a sequence of frame-wise binary masks
\begin{equation}
\mathcal{M}=\{m_t\}_{t=1}^{T}, \qquad m_t \in \{0,1\}^{H_t \times W_t},
\end{equation}
where $m_t$ denotes the segmentation mask of the target object in frame $I_t$, and $H_t$ and $W_t$ are the spatial height and width of the $t$-th frame, respectively.

Unlike conventional Ref-VOS formulations that implicitly assume the referred object is always present, we explicitly introduce an existence variable
\begin{equation}
e \in \{0,1\},
\end{equation}
where $e=1$ indicates that the referred target is present in the video, while $e=0$ denotes that the query does not correspond to any valid target instance. This variable is essential in practice, since large vision-language segmentation models may still produce plausible but incorrect masks even when the target is absent.

Accordingly, the overall prediction can be written as
\begin{equation}
\mathcal{M}=
\begin{cases}
\{\mathbf{0}\}_{t=1}^{T}, & e=0,\\
\Phi(V,q), & e=1,
\end{cases}
\end{equation}
where $\mathbf{0}$ denotes an all-zero mask and $\Phi(\cdot)$ represents the proposed multi-stage segmentation pipeline. In our framework, $\Phi$ is built upon a Sa2VA-centered semantic grounding module, followed by agent-guided anchor selection, mask propagation, and ambiguity-aware refinement.

\section{Method}
Our method follows a coarse-to-fine design that decouples semantic grounding, temporal propagation, and decision-level refinement. The key idea is to retain Sa2VA as the primary semantic interpreter of the language query, while introducing lightweight agentic modules to explicitly reason about target presence, anchor reliability, and frame-level ambiguity. The overall pipeline consists of five stages: (1) presence verification, (2) coarse Sa2VA segmentation, (3) anchor extraction from Sa2VA predictions, (4) SAM3-based propagation, and (5) planner-guided conflict resolution.

\subsection{Stage 1: Presence Agent}
The first stage is a \emph{Presence Agent}, whose sole purpose is to determine whether the target referred to by $q$ actually exists in the video. Formally, the agent predicts the binary decision
\begin{equation}
e = \Psi_{\mathrm{pres}}(V,q),
\end{equation}
where $\Psi_{\mathrm{pres}}$ denotes the presence judgment module.

If $e=0$, the system terminates early and outputs zero masks for all frames. This design prevents the downstream segmentation model from hallucinating object masks for non-existent targets, which is a common failure mode in open-ended language-conditioned video understanding. By separating existence reasoning from dense segmentation, the proposed framework reduces false positives and avoids unnecessary refinement on invalid queries.
\begin{table*}[t]
\centering
\small
\caption{Leaderboard results. Our method achieves third place on this challenge.}
\label{tab:leaderboard}
\begin{tabular}{c l c c c c c c}
\toprule
\# & Participant & J\&F & J & F & N-acc. & T-acc. & Final \\
\midrule
1 & HITsz\_Dragon & 0.7897 & 0.7680 & 0.8115 & 0.9615 & 0.9759 & 0.9091 \\
2 & goodx         & 0.7106 & 0.6880 & 0.7332 & 1.0000 & 0.9652 & 0.8919 \\
3 & \textbf{Ours}     & 0.7130 & 0.6882 & 0.7378 & 0.9615 & 0.9893 & 0.8879 \\
5 & junjie\_zheng & 0.6837 & 0.6633 & 0.7040 & 0.8846 & 0.9679 & 0.8454 \\
6 & rookie7777    & 0.6420 & 0.6145 & 0.6695 & 0.8462 & 0.9679 & 0.8187 \\
\bottomrule
\end{tabular}
\end{table*}
\subsection{Stage 2: Coarse Segmentation}
When the Presence Agent predicts $e=1$, we invoke Sa2VA on the full video and the referring expression to obtain an initial coarse segmentation trajectory:
\begin{equation}
\tilde{\mathcal{M}}
=\{\tilde{m}_t\}_{t=1}^{T}
=\mathrm{Sa2VA}(V,q).
\end{equation}
Here, $\tilde{m}_t$ denotes the coarse mask predicted for frame $I_t$. In our implementation, Sa2VA directly processes the entire video and returns frame-aligned predictions through \texttt{predict\_forward}, yielding a full-video segmentation hypothesis without requiring additional frame sampling or external temporal stitching.

We emphasize that the Sa2VA output is not treated as a disposable proposal. Instead, it serves as the semantic backbone of the entire pipeline. Since Sa2VA is responsible for grounding the referring expression to the correct visual entity, its outputs provide the most important semantic evidence about \emph{what} object should be segmented. However, due to limitations in mask precision and temporal stability, these coarse predictions are further refined by later stages rather than being used as the final output directly.

\subsection{Stage 3: Anchor Selection}
The third stage extracts reliable anchor frames from the Sa2VA trajectory. The purpose of this step is to transform Sa2VA's coarse semantic predictions into structured prompts that can initialize a stronger propagation model.

Let $\mathcal{A}\subseteq\{1,\dots,T\}$ denote the selected anchor set. For each anchor frame $a\in\mathcal{A}$, we derive geometric prompts from the corresponding Sa2VA mask $\tilde{m}_a$. In particular, we compute a bounding box
\begin{equation}
b_a=\mathrm{BBox}(\tilde{m}_a),
\end{equation}
and, if needed, additional point prompts extracted from the mask support region. These prompts are then passed to SAM3 as initialization signals.

This design ensures that SAM3 is not required to discover the target from scratch. Instead, it inherits object identity and coarse spatial localization from Sa2VA. In this sense, the two modules play complementary roles: Sa2VA determines \emph{which} object is relevant to the language query, while SAM3 improves \emph{how} this object is spatially localized and temporally propagated across the video.

\subsection{Stage 4: SAM3 Propagation}
Starting from the Sa2VA-derived anchors, we employ SAM3 to propagate the object masks over time and obtain a refined trajectory
\begin{equation}
\mathcal{T}=\{(\hat{m}_t,\hat{b}_t)\}_{t=1}^{T},
\end{equation}
where $\hat{m}_t$ and $\hat{b}_t$ denote the propagated mask and bounding box at frame $t$, respectively.

Compared with the raw Sa2VA predictions, this propagation stage mainly improves two aspects. First, it enhances temporal consistency by enforcing smoother object evolution across adjacent frames. Second, it sharpens spatial boundaries through stronger geometric mask decoding. As a result, SAM3 acts as a refinement and propagation engine that complements the semantic strengths of Sa2VA with improved localization fidelity.

\subsection{Stage 5: Planner-Based Selection on Ambiguous Frames}
Although SAM3 generally improves temporal smoothness and mask quality, it is not always optimal to fully replace the original Sa2VA predictions. In particular, when multiple frames are visually similar, when the target undergoes appearance changes, or when one source provides stronger semantic grounding while the other provides stronger geometric continuity, a frame-wise decision must be made between the two candidate results.

To address this issue, we introduce a lightweight \emph{Planner} that operates as a local decision module rather than a global controller. Specifically, for ambiguous frames, the Planner compares two candidate mask sources:
\begin{enumerate}[leftmargin=1.3em]
    \item the original Sa2VA prediction $\tilde{m}_t$;
    \item the propagated SAM3 prediction $\hat{m}_t$.
\end{enumerate}
Based on visual similarity, local temporal context, and prediction reliability, the Planner selects the more trustworthy candidate for each ambiguous frame. Therefore, its role is not to conduct exhaustive search over the video, but to resolve local conflicts between semantic grounding and geometric propagation.

Let $m_t^{*}$ denote the final selected mask at frame $t$. Then the final output sequence is written as
\begin{equation}
\mathcal{M}^{*}=\{m_t^{*}\}_{t=1}^{T},
\end{equation}
where each $m_t^{*}$ is chosen from either $\tilde{m}_t$ or $\hat{m}_t$ according to the Planner's decision rule. This final stage makes the pipeline robust to cases in which either Sa2VA or SAM3 alone would be suboptimal.

\subsection{Overall Perspective}
From a system perspective, the proposed framework is neither a pure Sa2VA pipeline nor a purely agent-driven architecture. Instead, it is an Agentic Ref-VOS framework in which semantic grounding is performed by Sa2VA, geometric propagation is handled by SAM3, and auxiliary agents provide explicit control over target existence, anchor reliability, and ambiguous-frame selection. This modular decomposition allows each component to specialize in the subproblem it is best suited for, thereby improving both robustness and segmentation quality in challenging referring video scenarios.

\section{Experiments}

Experiments are conducted on the MEVIS text split using the standard expression file \texttt{meta\_expressions.json}. We use Sa2VA-26B as the base Ref-VOS model and feed the full video together with the referring expression into Sa2VA. The Qwen3VL-32B initializes the Judger. The final result is conducted by  Selective Averaging~\cite{sasasa2va}.

\section{Conclusion}
This report presented an agentic Ref-VOS pipeline built around Sa2VA. The method begins with target presence judgment, uses Sa2VA to generate a coarse full-video mask trajectory, and then applies agentic planning, temporal search, verification, reflection, and SAM3 refinement to obtain the final output. The central viewpoint is simple: Sa2VA should provide the first grounded answer, but agents should decide how much to trust it, where to refine it, and when to reject it.

{
    \small
    \bibliographystyle{ieeenat_fullname}
    \bibliography{main}
}


\end{document}